\documentclass{INTERSPEECH2023}

\usepackage{amsmath}
\DeclareMathOperator*{\argmax}{arg\,max}

\usepackage{dirtytalk}
\usepackage{xcolor}

\usepackage[numbers]{natbib}

\setcitestyle{square}

\usepackage{sidecap}

\usepackage[T1]{fontenc}
\usepackage[utf8]{inputenc}

\usepackage{tipa}
\usepackage{svg}
\usepackage{caption}
\usepackage{multirow} 
\usepackage{graphicx}

\interspeechcameraready

\title{Probing self-supervised speech models for phonetic and phonemic information: a case study in aspiration}
\name{Kinan Martin, Jon Gauthier, Canaan Breiss, Roger Levy}
\address{
  Massachusetts Institute of Technology, USA}
\email{krmkrm@mit.edu, jon@gauthiers.net, canaan@mit.edu, rplevy@mit.edu}

\begin{document}

\maketitle

\begin{abstract}
Textless self-supervised speech models have grown in capabilities in recent years, but the nature of the linguistic information they encode has not yet been thoroughly examined. We evaluate the extent to which these models' learned representations align with basic representational distinctions made by humans, focusing on a set of phonetic (low-level) and phonemic (more abstract) contrasts instantiated in word-initial stops. We find that robust representations of both phonetic and phonemic distinctions emerge in early layers of these models' architectures, and are preserved in the principal components of deeper layer representations. Our analyses suggest two sources for this success: some can only be explained by the optimization of the models on speech data, while some can be attributed to these models' high-dimensional architectures.
Our findings show that speech-trained HuBERT derives a low-noise and low-dimensional subspace corresponding to abstract phonological distinctions.
\end{abstract}


\section{Introduction}

Self-supervised learning techniques have become the new standard for speech representation learning in recent years, and are at the foundation of models such as HuBERT \citep{hsu2021hubert} and wav2vec \citep{baevski2020wav2vec}, which are establishing new states of the art in automatic speech recognition \citep{baevski2020effectiveness,chung2020generative,liu2021tera}. These systems are pre-trained entirely on unlabeled data before being fine-tuned downstream for particular tasks. As such, they are free to derive whatever representations are optimal for their self-supervised pre-training task.

Recent work has asked whether the representations derived by these models are \emph{human-like} at multiple levels of granularity, from their specific representational contents \citep{pasad2022comparative,chung2021similarity} to their broad alignment with human brain responses to speech stimuli \citep{vaidya2022self,millet2022toward}.
Past work \citep{pasad2022comparative} shows that representations of speech stimuli extracted from self-supervised models can be successfully aligned with abstract phonetic descriptions of the input.

However, this prior work neglects a representational distinction crucial to linguistic theory: the difference between the \textbf{phonetic} level of representations and the \textbf{phonemic} level. Unlike phonetic representations, which are closely related to the acoustic features that implement them, phonemic representations function in linguistic theory as the axes of contrast which underpin lexical and grammatical distinctions and which are core to the effective use of language. Because these two levels are highly correlated in practice, distinguishing phonetic representational systems from phonemic ones is nontrivial. This paper explores whether self-supervised models encode distinct representations of speech inputs at both levels.

\subsection{Phones, phonemes, and allophones}

In the context of spoken languages, phonologists use the term \emph{phone} to describe the different phonetically-distinct categories that are targets of speakers' production and perception systems: for example, the \emph{b} in \textipa{[@"baUt]} \emph{about} and the \emph{p} in \textipa{["p\super{h}\ae{}t]} \emph{pat} are perceptually distinctive for a speaker English, and thus are classified as separate phones.\footnote{We give phonetic transcription in [square brackets] and phonemic transcription in /slashes/ with the International Phonetic Alphabet.}

However, there are many cases where multiple phones realize the same \emph{linguistically meaningful class} of sounds, the \emph{phoneme}. For example, the \emph{l} sounds in \textipa{[mI\textltilde{}k]} \emph{milk} and in \textipa{[lin]} \emph{lean} are perceptually distinguishable but not linguistically contrastive: there are no words that differ in their meaning based solely on whether the \emph{l} sound is \say{dark} \textipa{[\textltilde]} or \say{clear} \textipa{[l]}. These different phonetic realizations of the same phoneme are called \emph{allophones} \citep{chomsky1968sound,hayes2008introductory}.

\subsection{Probing models for phonetic and phonemic knowledge}
\label{sec:probe-design}

\begin{figure}[t]
  \centering
  \includegraphics[width=\linewidth]{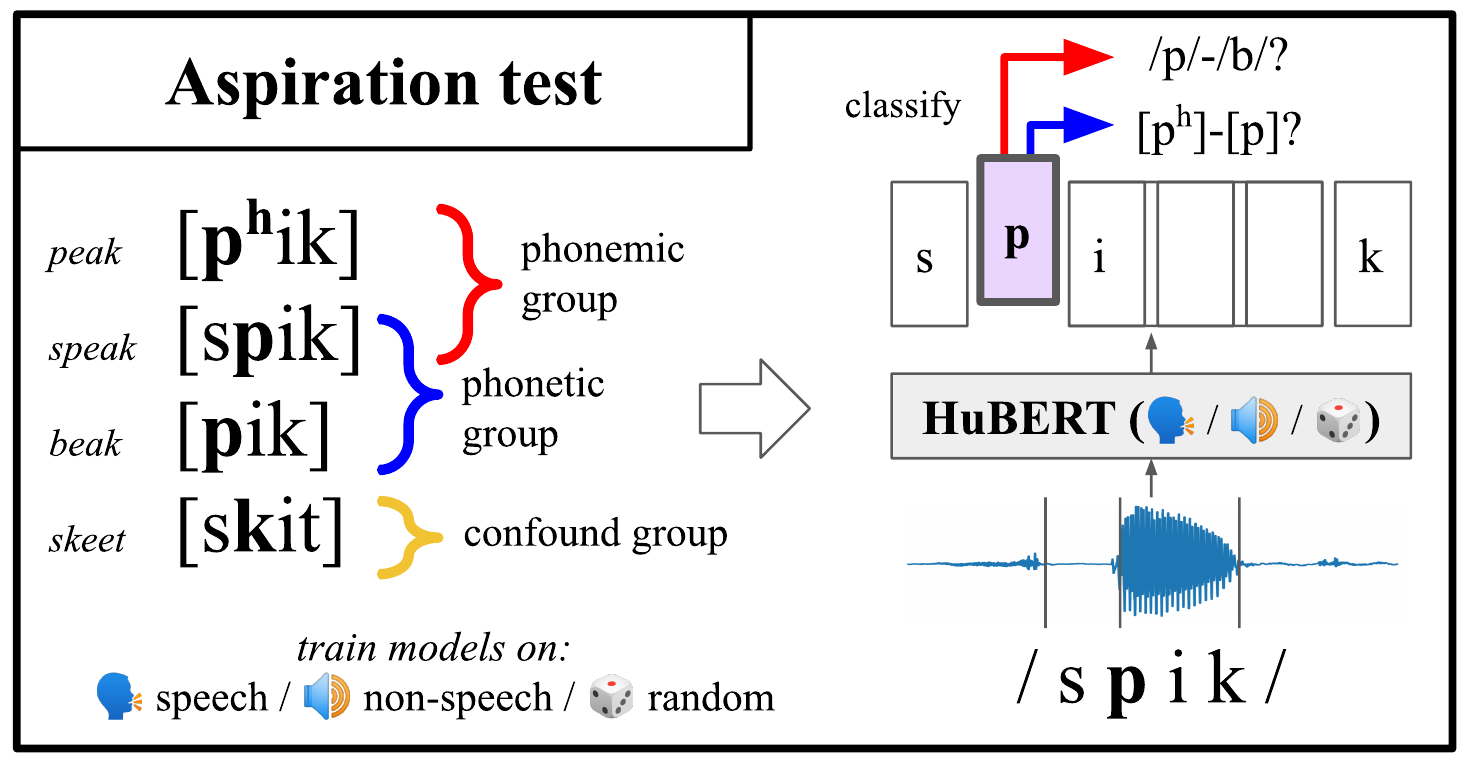}
  \vspace{-20pt}
  \caption{Diagram of classification probing paradigm.}
  \label{fig:classifier}
  \vspace{-15pt}
\end{figure}

\begin{SCtable*}
\scriptsize
\centering

\begin{tabular}{p{2.85cm}|lll}
\toprule
Contrast & Group 1 & Group 2 & Confound group \\
\midrule
Phonemic stop (ex. labial) & \#\textipa{\underline{p\super{h}}} V, \#\textipa{s\underline{p}} V & \#\textipa{\underline{b}} V & \#\textipa{s\underline{\{k,t,l,m,n,w\}}} V \\
Phonetic stop (ex. labial) & \#\textipa{\underline{p\super{h}}} V & \#\textipa{s\underline{p}} V, \#\textipa{\underline{b}} V & \#\textipa{s\underline{\{k,t,l,m,n,w\}}} V \\
\midrule
Consonant vs. vowel (+) & \underline{C} & \underline{V} & \emph{N/A}\\
Primary vs. no stress (+) & \underline{V1} & \underline{V0} & \emph{N/A}\\
Distant phoneme \mbox{    (before)} (-) & C X X X \underline{V} & V X X X \underline{V} & \emph{N/A}\\
Distant phoneme \mbox{    (after)} (-) & \underline{V} X X X C & \underline{V} X X X V & \emph{N/A}\\
\bottomrule
\end{tabular}%
\caption{Description of the stimuli entering into each contrast tested in our analyses. Columns denote targets of a classifier applied to input representations, and column contents denote IPA phonetic patterns used to select the relevant audio frames. \emph{\#} denotes word onset; \emph{C} a consonant; \emph{V} a vowel optionally followed by a stress number (\emph{1} = primary stress, \emph{0} = no stress)\citep{liberman1977stress}; \emph{X} any phoneme; \emph{\{\}} a set of options; \emph{(+)/(-)} positive/negative control tests. Representations were extracted from the onset to the offset of the \underline{underlined} phone in each matching audio file.}
\label{tbl:patterns}
\vspace{-20pt}
\end{SCtable*}

Although most phonemes do not share allophones, in cases of \emph{neutralization}, a specific phone could belong to one of two representationally-distinct phonemes. In this paper, we focus on the neutralization of post-sibilant and word-initial plosives in English. The top left panel of Figure~\ref{fig:classifier} demonstrates the relationship between the word forms \emph{peak}, \emph{speak}, and \emph{beak}, along with their phonemic and phonetic representations. In isolation, the phone \textipa{[p]} (voiceless, with short voice onset time) is ambiguous as to whether it represents phonemic /p/ (as in \textipa{[spik]} \emph{speak}) or word-initial phonemic /b/ (as in \textipa{[pik]} \emph{beak})  \citep{dmitrieva2015phonological}.

This paper tests whether self-supervised speech models derive distinctly \emph{phonemic} representations of their speech input, in addition to phonetic ones, focusing on the case of neutralization. Ultimately, we find that these speech models learn robust representations of phonetic and phonemic content, suggesting that these models learn human-like representations of acoustic input.  

\section{Methods}

We design a probing experiment \citep{belinkov-2022-probing,chung2021similarity} to test whether self-supervised speech model representations contain distinct phonetic and phonemic contents. We use the Massive Auditory Lexical Decision (MALD) database \citep{tucker2019massive}, which contains recordings of 26,793 words and 9,592 pseudowords, each uttered in isolation by a single speaker, accompanied by time-aligned phonemic annotation.
We convert the recordings to mono with a 16 kHz sample rate, and present them to a suite of self-supervised speech models, each pre-trained on different datasets.



\subsection{Speech models and representations}
\label{sec:models}

Our tests evaluate the representations of HuBERT \citep{hsu2021hubert}, a popular self-supervised speech representation architecture. This model has been widely used as a component of larger spoken language processing pipelines and has proven useful for downstream language tasks \citep{baevski2020effectiveness,Wu2020,nguyen2020investigating,Ravi2020}. It has also been analyzed in other works probing self-supervised speech models \cite{pasad2022comparative,chung2021similarity,millet2022toward}.

The HuBERT architecture\footnote{We analyze the HuBERT \say{Base} model in this paper, which has approximately 95 million parameters.} is composed of a seven-layer convolutional neural network (CNN) which feeds into a twelve-layer transformer. The CNN takes in raw audio at 16 kHz and yields a sequence of frames, each 20 milliseconds long (50 Hz). These frames are then fed through a stack of 12 transformer layers. In our work, we extract the representations output by each of the CNN layers and transformer blocks.\footnote{HuBERT also includes a final classification layer which predicts the identity of masked input frames. These classification weights and the discrete output codes are not studied in this paper.}

We evaluated instances of the HuBERT architecture trained on three different objectives. These objectives were selected to help us distinguish the specific effects of training on speech data:
\begin{itemize}
    \item \textbf{speech HuBERT}: trained to predict masked frames on speech data from LibriSpeech \citep{panayotov2015librispeech}, as released by \citep{hsu2021hubert}
    \item \textbf{non-speech HuBERT}: trained on non-speech audio from the AudioSet dataset \citep{gemmeke2017audio}, as released by \citep{millet2022self}
    \item \textbf{random HuBERT}: a matching architecture with randomly initialized CNN and transformer weights
\end{itemize}
The non-speech and random models were included to examine the extent to which the audio-based non-speech training objective and architecture alone facilitate the learning of phonological representations, respectively.

We also analyzed a log-mel representation of the acoustic input, which computes the log-power of acoustic input within 80 mel frequency bands as computed by Kaldi's \texttt{Fbank} routine. This model targets low-level features of the acoustic signal, acting as a suitable baseline of comparison for the more complex HuBERT-based speech models.


HuBERT produces a representation $h_\ell(t)$ for each 20ms frame of the audio input at each layer $\ell$. We first extract frame-level representations $h_\ell(t)$ for each file in the MALD dataset and for each HuBERT model instance.
Because phones typically span more than one frame, we take a model's representation of a phone to be its mean activation over the overlapping frames $h_\ell(t)$ \citep{pasad2021layer}. This aggregation produces a vector of size $d$ for each phone in the dataset, which is the object of our analyses.



\subsection{Phonetic and phonemic probes}
\label{sec:probe-impl}

We designed a set of classification tasks exploiting the pattern of phonetic neutralization of the stop-voicing contrast in English, described in Section~\ref{sec:probe-design}. We implemented these tasks for triples of phones at three places of articulation: labial (/p/, /sp/, /b/), alveolar (/t/, /st/, /d/), and velar (/k/, /sk/, /g/). Table~\ref{tbl:patterns} demonstrates this scheme for the labial place of articulation, which we use for exposition in the main text. Quantitative results are the average of identical tasks at three places of articulation. 

For each place of articulation, we design \emph{phonemic} and \emph{phonetic} multinomial regression probes $p_m$ and $p_t$. Given a representation of a particular phone at hidden layer $h_\ell$, our probes learn classifier weights
\begin{equation}
    p_m(h_\ell) \propto \exp(W_m^T h_\ell);\qquad p_t(h_\ell) \propto \exp(W_t^T h_\ell)
\end{equation}

The phonemic classifier $p_m$ is tasked with distinguishing phonemic /p/ from phonemic /b/ as realized in different contexts (Table~\ref{tbl:patterns}). In contrast, the phonetic classifier $p_t$ must distinguish cases of aspirated \textipa{[p\super{h}]} from cases of unaspirated \textipa{[p]} and \textipa{[b]}.

\subsubsection{Controlling for phonetic confounds}
\label{sec:sconjecture}

These phonetic and phonemic labelings of the data are confounded with a simpler disjunctive phonetic coding of the input: our phonemic labial contrast (shown in the first row of Table~\ref{tbl:patterns}) could be solved by grouping together inputs which contain \textipa{[p\super{h}]} or which are directly preceded by the phone [s].\footnote{This information is likely to be present in the model representations due to either 1) co-articulation of [s] and [p] in the input, 2) overlapping spectral signals of [s] in the earliest input frames near the onset of [p], or 3) models' combination of neighboring low-level acoustic features during their feed-forward pass.} This means a model could solve the phonemic contrast by searching for the presence of a word-initial [s], rather than attending to the neutralized stop itself. 

To control for this, we design a third class of \textbf{confound} inputs, shown in the third column of Table~\ref{tbl:patterns}. Phones in this class have the property that they are also directly preceded by the sound [s] and followed by a vowel. We constrain our probe classifiers to jointly contrast among these three classes, thus ruling out a classification strategy which merely exploits the presence of [s] to solve the phonemic contrast.

Our classification setup thus learns three-way multi-class weights $W_m, W_t \in \mathbb R^{3 \times d}$ for a given $d$-dimensional input representation. We optimize these weights under a multinomial loss, requiring the classifier to jointly contrast inputs between the two classes of interest and also with the confound class.
We then evaluate these classifiers on held-out data with a multi-class ROC/AUC metric, computing an ROC score for each of the three possible binary contrasts in the data, weighting each score by the prevalence of the classes in the contrast, and summing these weighted scores.

\subsection{Dimensionality reduction}
\label{sec:dim-reduction}

Previous work has suggested that HuBERT encodes some information at much higher representational levels, such as word identity and meaning \citep{pasad2021layer}. In order to avoid confounding our phonetic and phonemic tests with these higher levels, we designed a set of \textbf{control tests} to help us target a distinctly phonological level of representation within the models. Our positive and negative control tests, shown in the lower section of Table~\ref{tbl:patterns}, were designed to select for a maximal and minimal level of representational capacity. These tests offer a non-circular method for selecting views of model representations which are neither too strong (i.e., performing above chance on negative controls) nor too weak (i.e., performing below ceiling on positives).



For each model representation $h$, we use these control tests to search for a representation of reduced dimensionality $h^d$ which satisfies the above criteria.
Concretely, let $h_\ell^d: \mathbb R^{n \times d}$ denote a model's representations of all $n$ phones in the dataset at layer $\ell$ reduced to $d$ principal components. Let $P_1, P_2$ and $N_1, N_2$ denote ROC/AUC measures of held-out probing performance on a representation $h_\ell^d$ for the positive and negative controls respectively, shown in Table~\ref{tbl:patterns}. We define a \emph{control score} $\mathcal{S}$ summarizing across-layer difference at dimension $d$:
\begin{align}
    \mathcal{S}^d(h) = \sum_\ell & \left( P_1(h_\ell^d) + P_2(h_\ell^d) - N_1(h_\ell^d) - N_2(h_\ell^d) \right)
\end{align}
For each model, we selected a \textbf{constrained} dimensionality $d^*$ which maximized this contrast between positive and negative controls:
\begin{equation}
    d^*(h) = \argmax_{d \in \{2^1, 2^2, \dots, D\}} \mathcal{S}^d(h)
\end{equation}

\section{Results}

\begin{figure}[t]
  \centering
  \includegraphics[width=\linewidth]{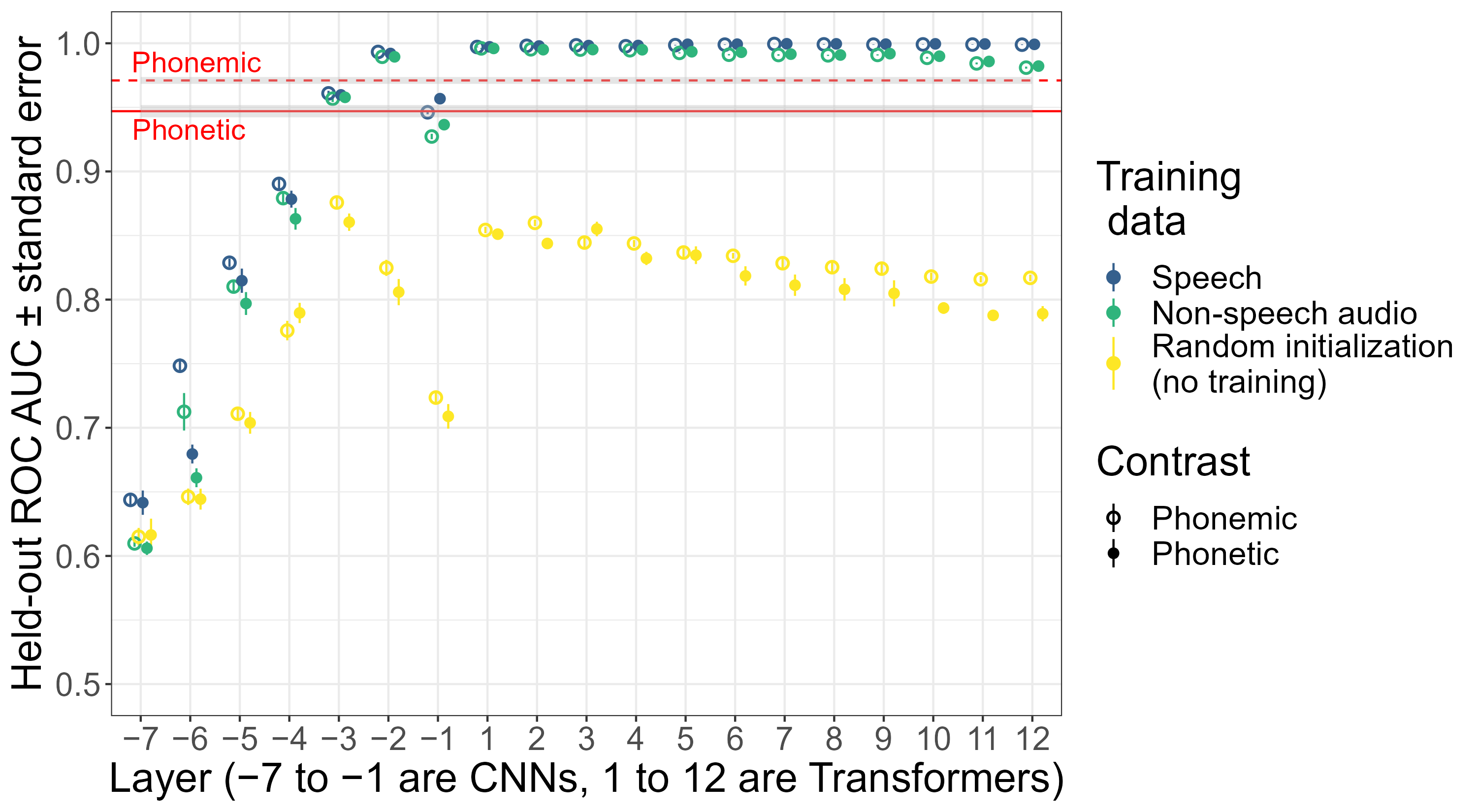}
  \caption{Layer-wise performance of HuBERT models on phonetic and phonemic classifiers; log-mel classifier baselines in red.}
  \label{fig:asipration_test}
\vspace{-15pt}
\end{figure}

We train and evaluate phonetic and phonemic probes for representations extracted from every HuBERT layer (7 CNN layers and 12 transformer layers) optimized for each training objective (speech, non-speech audio, or randomly initialized weights). We report the probe performance as the weighted ROC/AUC metric described in Section~\ref{sec:sconjecture}. We estimate this metric by nested 10-fold cross validation, using an inner cross-validation loop to estimate an L2 regularization hyperparameter.

\subsection{Aspiration test, original dimensionality}

Figure \ref{fig:asipration_test} shows classifier performance on the aspiration test conducted on the HuBERT speech, non-speech audio, and random models, along with a log-mel baseline. Probe performance is evaluated with an ROC/AUC metric, where 0.5 corresponds to random chance guessing and 1.0 corresponds to perfect phonetic or phonemic contrasts. Since the log-mel metric is external to the models, we show it as a horizontal red line across layers.

In all models, we see a rapid transition from near-chance performance to high scores on both phonetic and phonemic contrasts within the first several CNN layers. Each model then exhibits a performance drop at the final layers of the CNN ($x$$=$$-1$ in Figure~\ref{fig:asipration_test}), followed by near-ceiling performance in the task-optimized transformer models and far worse performance in the randomly initialized transformer layers.
We see a slight decrease in probe performance in the final layers of the non-speech audio model, suggesting task-specific specialization away from speech features.

The random-weights model performs worse than task-optimized models but better than chance, demonstrating that some nontrivial portion of success on these tests can be attributed simply to high-dimensional random projection. However, we see that both task-optimized HuBERT models reach ceiling performance at intermediate layers, suggesting that training selects for quality phonetic and phonemic representations.

These results show that the speech models' representations are both phonetically and phonemically robust, and that these abstract representations are acquired early in the feed-forward pass in the CNN, even before the first transformer block.
However, this at-ceiling performance may mask meaningful differences between models and contrasts of interest. Using the dimensionality reduction tools described in Section~\ref{sec:dim-reduction}, we asked if models continue to encode these abstract distinctions in their highest-order principal components.


\subsection{Control tests, constrained dimensionality}
\label{sec:control_test_results}

\begin{figure}[t]
  \centering
  \includegraphics[width=\linewidth]{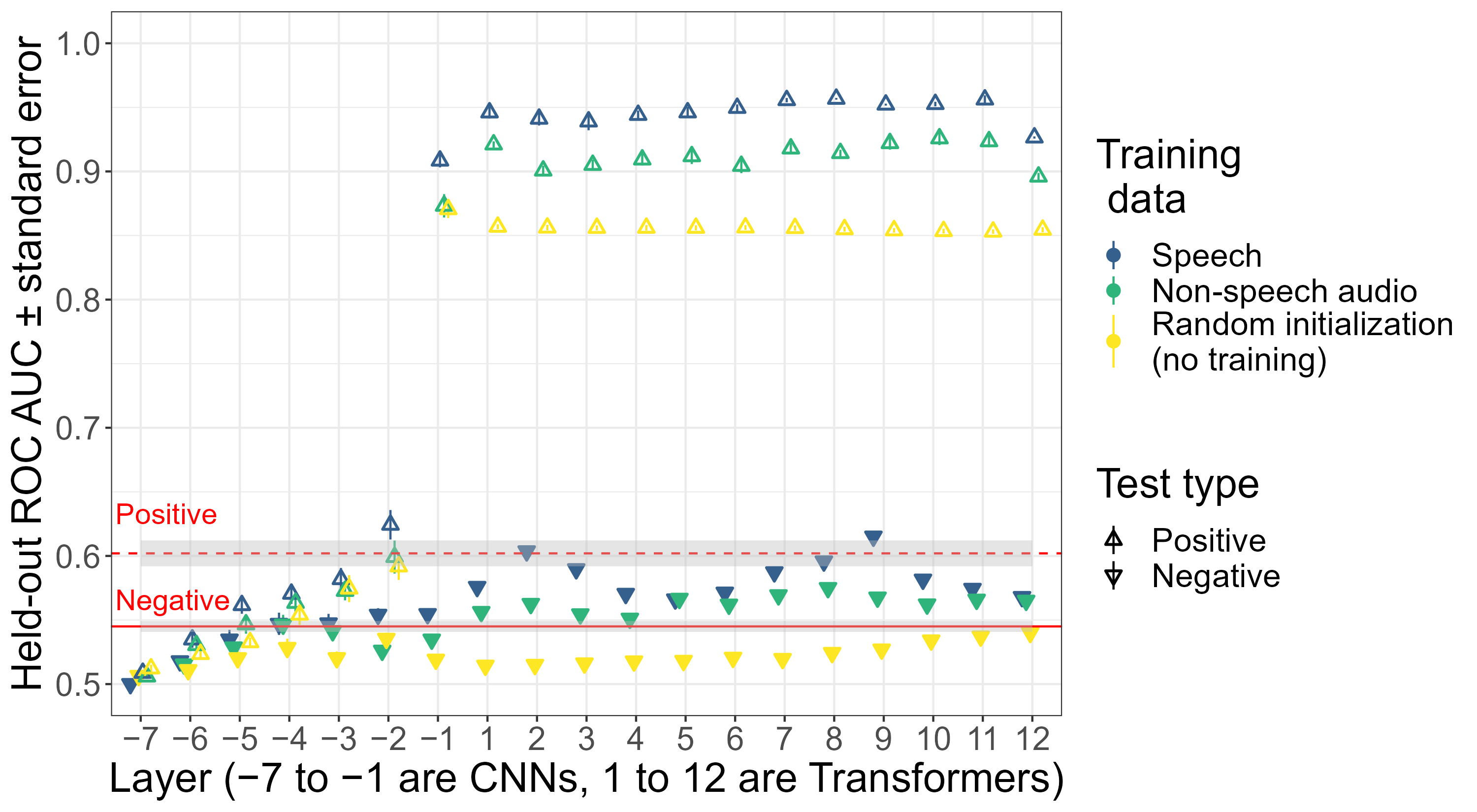}
  \caption{Layer-wise performance of HuBERT models on control tests on constrained dimensionality; log-mel classifier baselines in red.}
  \label{fig:controls_optimal_d}
\vspace{-15pt}
\end{figure}


Figure \ref{fig:controls_optimal_d} shows the layer-wise performance of the classifier on the positive and negative control tests conducted on the HuBERT speech, non-speech audio, and random models at a constrained dimensionality as described in Section~\ref{sec:dim-reduction}. Speech and random models: $d^*$$=$$16$; non-speech audio model: $d^*$$=$$8$, log-mel: $d^*$$=$$80$ (no reduction).

Although our probe performed at ceiling in the aspiration test, the soundness of our probing paradigm is affirmed by the patterning of performance on our control tests: all HuBERT models perform poorly at the negative controls and well at the positive controls, confirming that there do exist non-phonological distinctions that the models do not encode.

For all models, a marked jump in performance occurs in the last layer of the CNN. The layer at which this jump occurs coincides with the layer at which model performance temporarily drops in the aspiration test. This suggests that the sudden improvement in representing gross phonological categories (the positive controls) trades off with success at fine-grained phonological distinctions (the aspiration test), though this performance picks back up later in the transformer layers (Figure 4).

\subsection{Aspiration test, constrained dimensionality}

\begin{figure}[t]
  \centering
  \includegraphics[width=\linewidth]{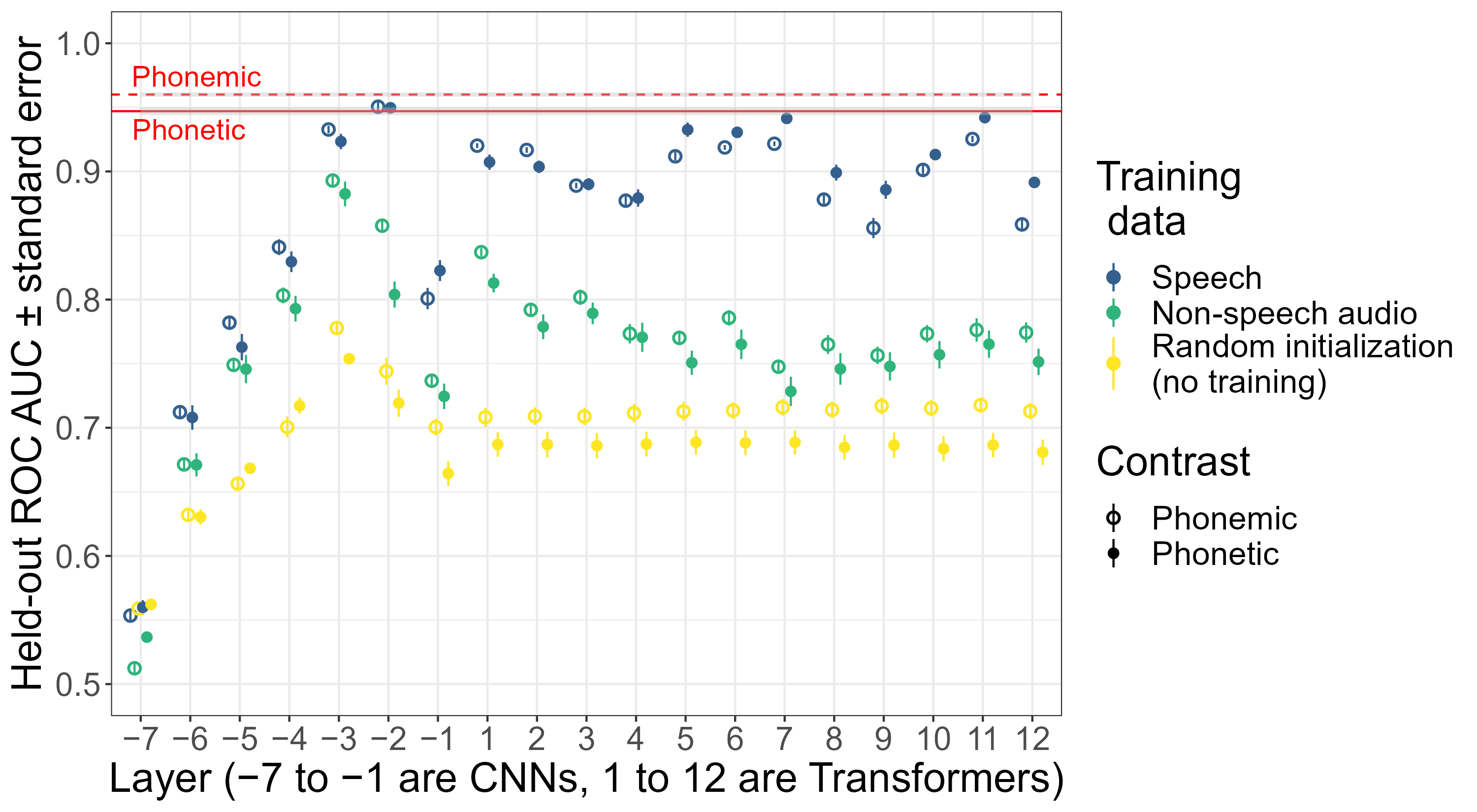}
  \caption{Layer-wise performance of HuBERT models on phonetic and phonemic classifiers with constrained $d^*$; log-mel baselines in red.}
  \label{fig:aspiration_optimal_d}
\vspace{-15pt}
\end{figure}

Figure~\ref{fig:aspiration_optimal_d} shows classifier performance on the aspiration test conducted at the constrained dimensionalities $d^*$. Here, the models no longer perform at ceiling on the classification task. The performance of log-mel remains the same as in Figure~\ref{fig:asipration_test} since its constrained $d^*$ value is unreduced, remaining at $d^*$$=$$80$. Unlike the HuBERT models, log-mel is unable to perform well at the positive controls (visible in Section~\ref{sec:control_test_results}), meaning that there is no downward pressure when finding a constrained $d^*$ that maximizes the difference between positive and negative controls.
The HuBERT speech model out-performs both other models, and in both late CNN layers and early transformer layers achieves better performance on phonemic classification than phonetic. In later layers, however, the performance of phonetic distinctions gradually improves relative to the phonemic ones. This suggests that, while the self-supervised training objective leads to the HuBERT speech model discovering both salient phonetic and phonemic abstractions, the phonetic representation may be prioritized in later layers due to its greater utility in predicting the identity of short (20ms) masked frames in training.

\vspace{-5pt}
\section{Discussion}
\label{sec:discussion}

We found in both aspiration and control tests that HuBERT successfully draws abstract representational distinctions between input phonemes within its CNN layers, prior to even the first transformer block. This suggests that the more complex transformer architecture may actually be overpowered for these types of speech representation tasks. This is compatible with some work on distillations of HuBERT, which find that a majority of HuBERT's transformer layers can be excised without significant performance loss, so long as the CNN layers are maintained \citep{sanh2019distilbert}. Future work can investigate the degree to which these phonetic and phonemic distinctions are maintained in distilled models, and whether models with fewer parameters could recover the same representational contents from scratch.

We find both phonetic and phonemic distinctions are encoded early in the HuBERT speech model, as well as to a lesser degree in the non-speech audio model. Our results demonstrate that HuBERT's forward pass first recapitulates, then exceeds, the representational capacity of the log-mel baseline. First, the penultimate CNN layer of HuBERT trained on speech data produces representations which encode the same phonological distinctions as log-mel features (Figure 4, $x$$=$$-2$). Next, the final CNN layer of HuBERT renders higher-level phonological distinctions not present in the log-mel features (Figure 3, $x$$=$$-1$).

Some of this success seems to be due merely to HuBERT's high-dimensional transformer architecture. We note that even the random-weights model encodes coarse-grained phonological distinctions late in its CNN layers (Figure 3, $x$$=$$-1$, green). However, the fine-grained distinctions targeted in the aspiration test are not encoded at the same layers (Figure 2, $x$$=$$-2$ and $-1$, green). This disparity persists through the transformer layers, where we see a sustained gap in decoding performance between the two models (Figures 2 and 3, blue vs. green).

The contrasts with the above baseline models suggest that task-optimized HuBERT simultaneously and with lower noise accomplishes two representational functions modeled independently by the log-mel and random projection baselines: it derives the fine-grained phonemic distinctions readable from log-mel representations, while rendering decodable the gross phonological distinctions present in the randomly initialized models, all without losing lower-level phonetic information.

Although we sought to control for frame overlap confounds and select only information relevant to the phonological identification task, confounds may remain. In the aspiration test, the non-speech and random models unexpectedly performed better at phonemic classification than the phonetic one. We expected the opposite result, given \cite{dmitrieva2015phonological}'s finding that the phonetic /sp/-/b/ similarity can indeed be recovered from low-level acoustic features. Further, the constrained speech model unexpectedly gains higher relative performance on the phonetic than the phonemic task over the course of its transformer layers (Figure 4, blue).

A possible explanation for this discrepancy is that the MALD dataset's speaker exhibits idiolectic differences in voice onset time that undermine the assumed acoustic similarity of /sp/ and word-initial /b/, which lie outside the generalization population of \cite{dmitrieva2015phonological}'s claim.
This would lead to degradation on the phonetic classification task relative to the phonemic one, since the \say{unaspirated} category would then be internally heterogeneous. Extending the same analysis to a multi-speaker dataset or creating new phonological tests may resolve this issue.



\vspace{-5pt}
\section{Conclusion}

This paper tested whether self-supervised speech models derive distinctly phonemic representations of their speech input, using aspiration as a case study. We find that these speech models derive robust representations of phonemic and phonetic content which recover and surpass fine-grained log-mel features. These representations emerge early in the models' processing stream, a phenomenon which has implications for the design of self-supervised speech models. Differences in model performance depending on the content of training datasets are evidence for task specialization arising in later model layers. Overall, the lack of marked distinction between phonetic and phonemic contrasts suggests the presence of certain confounds that complicate representational probing of these speech models.

{\footnotesize\noindent\textbf{Acknowledgments: }Thanks to Juliette Millet and Ewan Dunbar for graciously providing their non-speech audio-trained models, to the MIT Exp/Comp group for feedback. KM, CB, and JG gratefully acknowledge support from MIT CPL, the MIT--IBM Watson AI Lab, and the Open Philanthropy Project, respectively.}

\section{References}

\bibliographystyle{IEEEtran}
\bibliography{mybib}

\end{document}